\pdfoutput=1
\documentclass[11pt]{article}

\usepackage[]{acl}

\usepackage{times}
\usepackage{latexsym}

\usepackage[T1]{fontenc}
\usepackage[utf8]{inputenc}
\usepackage{amsfonts, amsmath, amssymb,mathtools,bbm,bm}
\usepackage{centernot}
\usepackage{microtype}
\newcommand{\E}{\mathbb{E}}
\newcommand{\tr}{\text{Tr}}

\usepackage{amsthm}
\theoremstyle{plain}

\newtheorem{lemma}{Lemma}

\newtheorem{property}{Property}

\usepackage[shortcuts]{extdash}  % make hyphen \-/
\usepackage{hyperref}

\title{Latent Positional Information is in the Self-Attention Variance of \\Transformer Language Models Without Positional Embeddings}

\author{Ta-Chung Chi$^\dagger$ \\
  Carnegie Mellon University \\ \And
  Ting-Han Fan \\
  Princeton University \\\And
  Li-Wei Chen \\
  Carnegie Mellon University \\\AND
  Alexander I. Rudnicky \\
  Carnegie Mellon University \\\And
  Peter J. Ramadge \\
  Princeton University \\
}
\begin{document}
\maketitle
\begingroup\def\thefootnote{$\dagger$}\footnotetext{{Correspondence to: \texttt{tachungc@andrew.cmu.edu}}}\endgroup
\begin{abstract}
The use of positional embeddings in transformer language models is widely accepted. However, recent research has called into question the necessity of such embeddings. We further extend this inquiry by demonstrating that a randomly initialized and frozen transformer language model, devoid of positional embeddings, inherently encodes strong positional information through the shrinkage of self-attention variance. To quantify this variance, we derive the underlying distribution of each step within a transformer layer. Through empirical validation using a fully pretrained model, we show that the variance shrinkage effect still persists after extensive gradient updates. Our findings serve to justify the decision to discard positional embeddings and thus facilitate more efficient pretraining of transformer language models.
\end{abstract}

\section{Introduction \& Related Work}
Transformer models have become the backbone of natural language processing applications~\cite{vaswani2017attention,bert,gpt}. Within the transformer architecture, there are two main categories: 1) bidirectional models, such as BERT~\cite{bert}, that are trained using the masked language modeling objective, and 2) (causal) language models, such as GPT~\cite{gpt}, that are trained using the traditional language modeling objective. Both of these categories share the common feature of using positional embeddings for encoding token distance.

Whether positional embeddings are truly essential has been a subject of ongoing research. While they have been considered necessary for bidirectional transformer models~\cite{pmlr-v97-lee19d,luo-etal-2021-positional,sinha-etal-2021-masked,haviv2022transformer}, the situation is different for transformer language models~\cite{Irie2019LanguageMW,yang-etal-2019-assessing,tsai-etal-2019-transformer,scao2022what,haviv2022transformer}. In transformer language models, the removal of positional embeddings results in only a marginal decline in performance, while enabling more efficient training~\cite{haviv2022transformer}.
In addition to empirical evidence, it has been proven~\cite{bhattamishra-etal-2020-computational} that transformer language models without positional embeddings are Turing-complete and able to model sequences akin to recurrent neural networks~\cite{6302929,osti_6910294}. Despite this, it remains an open question where positional information is stored in the absence of positional embeddings. This motivates further investigation into individual operations within a transformer layer.

\begin{figure}
    \centering
    \includegraphics[width=\linewidth]{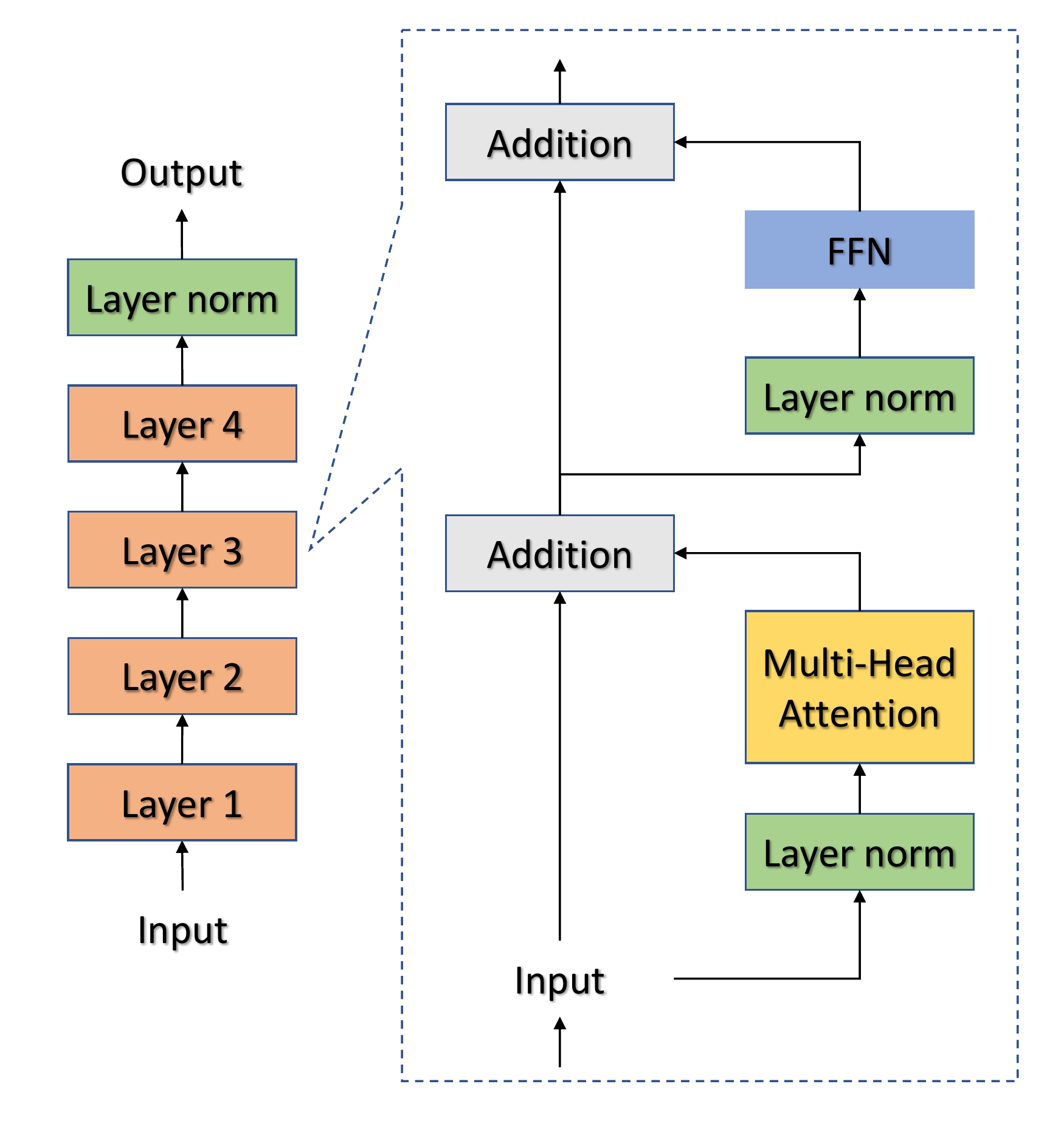}
    \caption{The architecture of a Pre-LN transformer language model. All the parameters are randomly initialized and randomly sampled input is used in this work.}
    \label{fig:pre_ln}
\end{figure}

The example architecture of a pre-LN~\cite{Xiong2020OnLN} multi-layer transformer language model with no positional embeddings used in this work is shown in Figure~\ref{fig:pre_ln}.\footnote{Post-LN places layer norm at different positions. It is the configuration used in BERT~\cite{bert}.}
We hereinafter refer to this configuration as TLM.
Our primary focus is on the multi-head attention (MHA) module of a randomly initialized TLM, as it is the only module that allows inter-token information exchange. To gain a deeper understanding, we compute the mean and variance of MHA outputs. To our surprise, we discover that the variance already encodes latent positional information, with later tokens in a sequence displaying smaller variance.
This motivates us to quantify the variance by deriving the output distribution after MHA operations. Finally, through empirical validation using a fully pre-trained TLM, we confirm thatthe same variance shrinkage effect persists after extensive gradient updates.

To the best of our knowledge, we are the first to identify and quantify the latent positional information in TLMs. Our results provide theoretical insights into the removal of positional embeddings, enabling more efficient pretraining of future TLMs.

\begin{figure}
    \centering
    \includegraphics[width=\linewidth]{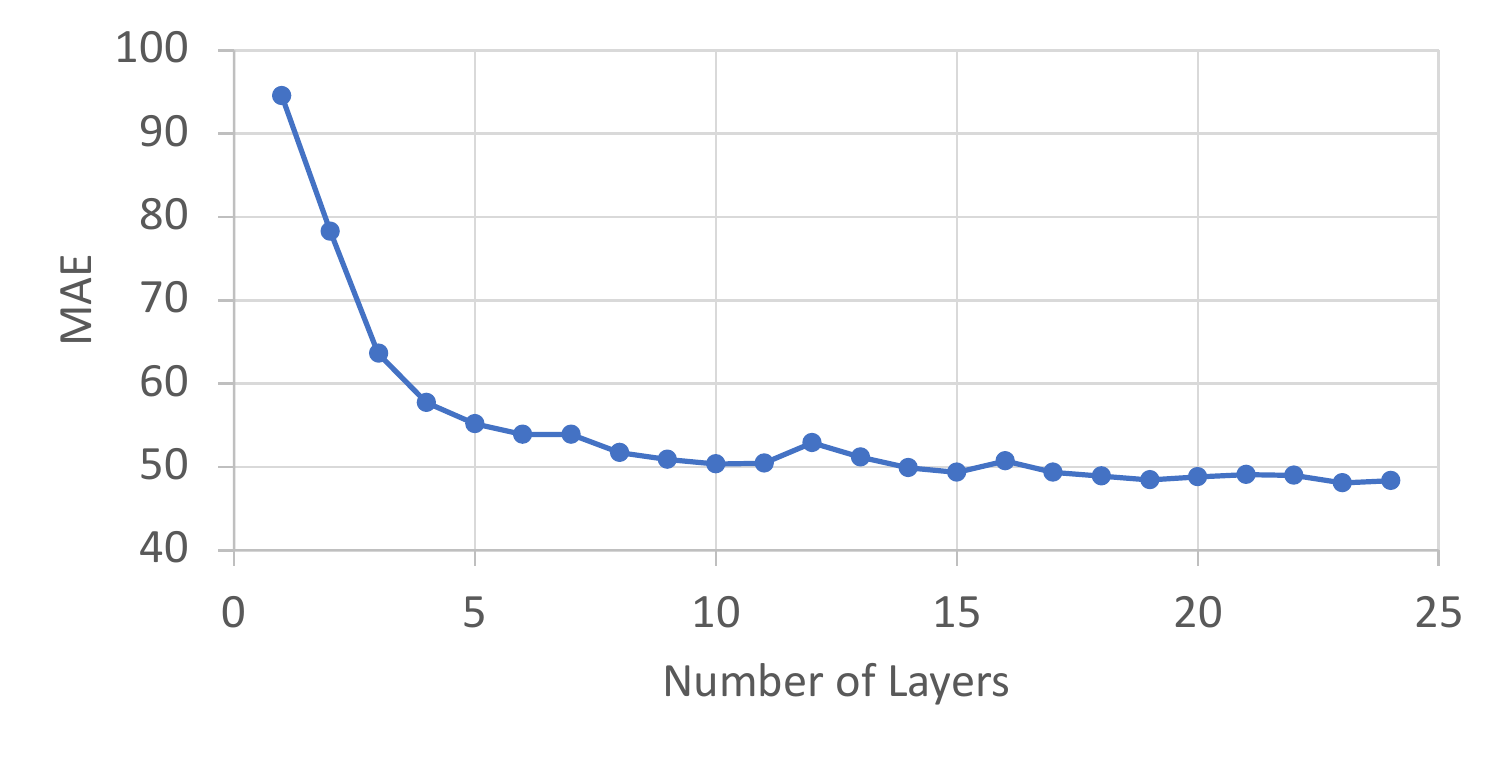}
    \caption{We plot the positions w.r.t their mean absolute error (MAE) for input sequence length $L=512$. A naive baseline of predicting the middle point of $L=256$ gives an MAE of $128$. The numbers are the average of 5 seeds.}
    \label{fig:mae}
\end{figure}

\section{Probing Experiments}
Given BERT and TLM (GPT) with positional embeddings removed, prior work~\cite{haviv2022transformer} shows that only TLM is able to maintain the same language modeling performance as its original version with positional embeddings.
The discrepancy might be explained by the fact that 
only TLM encodes positional information within its layers, as shown by the position probing experiment in~\citet{haviv2022transformer}. Since both BERT and TLM have access to the same semantic input and the only difference is the use of causal attention masks in TLM, we hypothesize that the positional information may be attributed to the interaction between causal attention masks and the TLM architecture.

To further explore this hypothesis, we use a randomly initialized and frozen TLM to eliminate any semantic influence and focus solely on the architectural design. Additionally, to prevent the model from memorizing the order of input sequences, we do not perform embedding lookups and feed the model with randomly sampled input vectors. A trainable two-layer linear classifier with ReLU activation in between was appended to the TLM to probe the position of each token (further details can be found in Appendix~\ref{sec:appendix_probing}). We plot the mean absolute error (MAE) w.r.t the number of transformer layers in Figure~\ref{fig:mae}. The plot indicates a randomly initialized and frozen TLM with randomly sampled input vectors inherently provides positional information, with an increase in the number of layers resulting in higher probing performance. This surprising outcome prompts further investigation into the encoding of latent positional information inside the TLM architecture.

\section{Theoretical Analysis}
\label{sec:analysis}
We dissect the inner workings of a TLM by deriving the distribution of TLM operations in the hope that they elucidate where the latent positional information is stored.
The derivation is made possible thanks to the usage of a randomly initialized and frozen TLM. We adopt the initialization settings in accordance with those employed in GPT~\cite{gpt}. WLOG, our derivation is limited to the operations of the first layer in a TLM and the FFN component is omitted (justified in \S\ref{sec:final_ln}). The hyperparameters utilized in the simulations are: hidden dimension $d=768$, number of attention heads $H=12$, head dimension $d/H=64$, sequence length $L=512$, standard deviation for initialization $\sigma=0.02$. All proofs of lemmas are deferred to Appendix~\ref{sec:appendix_proof}.

Given a sequence of randomly sampled input embeddings $\{\bm x_m\}_{m=1}^L$, where each element of $\bm x_m\in\mathbb{R}^d$ is sampled i.i.d from $N(0, \sigma^2)$, a TLM consists of the following operations:

\subsection{Layer Normalization}
\label{sec:ln}
For each input embedding $\bm x_m$, it computes the sample mean and (biased) sample variance:
\begin{equation*}
    \overline{\bm x}_{m,:}=\frac{\sum_{i=1}^d \bm x_{mi}}{d},~S(\bm x_{m,:})=\frac{\sum_{i=1}^d (\bm x_{mi}-\overline{\bm x}_{m,:})^2}{d}
\end{equation*}
Then each entry $i$ of $\bm x_m$, denoted as $\bm x_{mi}$, is normalized by mean and variance to $\bm e_{mi}$:
\begin{align*}
    \bm e_{mi}&=\frac{\bm x_{mi} - \overline{\bm x}_{m,:}}{\sqrt{S(\bm x_{m,:})}}*\gamma+\beta \\
    &\overset{(*)}{\approx}\frac{\bm x_{mi} - \mathbb{E}[\bm x_{mi}]}{\sqrt{\mathbb{V}[\bm x_{mi}]}}\sim N(0,1),
\end{align*}
where $\mathbb{V}[\bm x]$ denotes the variance of $\bm x$. Since the initialization scheme sets $\gamma=1$ and $\beta=0$, $(*)$ holds with sufficiently large $d$ by the Law of large numbers and the continuous mapping theorem.

\begin{figure}
    \centering
    \includegraphics[width=\linewidth]{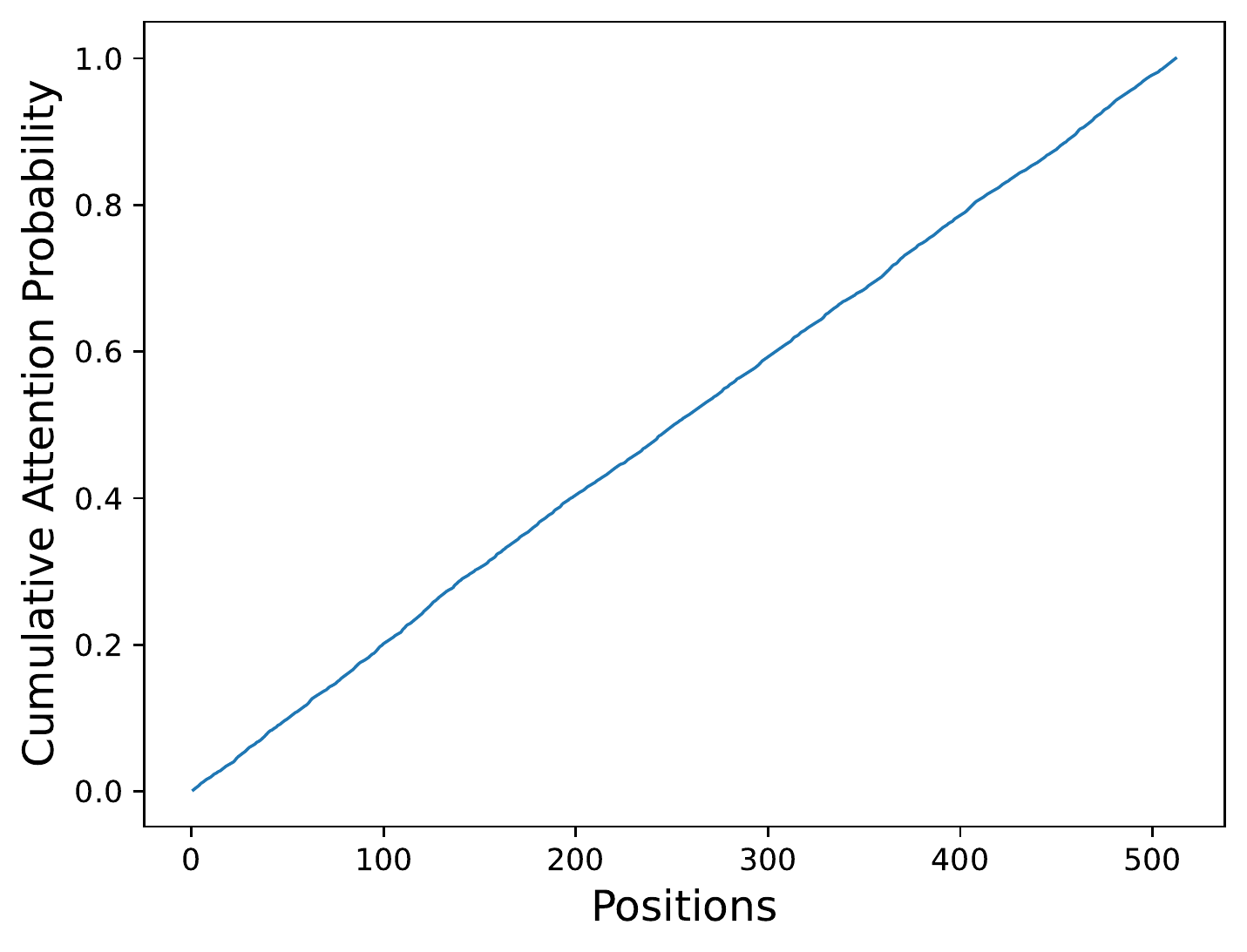}
    \caption{We plot the positions w.r.t their cumulative attention score for $L=512$ averaged over 500 samples.}
    \label{fig:avg_attn}
\end{figure}

\subsection{Self Attention}
\label{sec:lemmas-for-attn}
Each attention head computes query, key, and value vectors in $\mathbb{R}^{\frac{d}{H}}$:
\begin{align*}
    \bm q_m=\bm W_q\bm e_m,\enskip\bm k_m=\bm W_k\bm e_m,\enskip\bm v_m=\bm W_v\bm e_m,
\end{align*}
where $\bm W_q$, $\bm W_k$, $\bm W_v\in\mathbb{R}^{\frac{d}{H}\times d}$ are matrices with each element sampled i.i.d from $N(0,\sigma^2)$.

To be precise, most matrices ($\bm W_q^{(h)}$, $\bm W_k^{(h)}$, $\bm W_v^{(h)}$), vectors ($\bm q_m^{(h)}$, $\bm k_m^{(h)}$, $\bm v_m^{(h)}$), and scalars ($l_{mn}^{(h)}$, $a_{mn}^{(h)}$) are associated with a head number $h$. For notation simplicity, we only show the dependency on $h$ when we need it.

\begin{lemma}
    $\bm q_m$, $\bm k_m$, and $\bm v_m$ have zero mean and $(d\sigma^2)\cdot I$  covariance matrix.
    \label{lemma:cov_v}
\end{lemma}
\noindent The resulting vectors are processed by the self-attention module for pre-Softmax logits:
\begin{equation*}
    l_{mn}= 
\begin{cases}
    \langle \bm q_m, \bm k_n \rangle,& \text{if } m\geq n\\
    -\inf,              & \text{otherwise}
\end{cases}
\end{equation*}
followed by the scaled softmax normalization:
\begin{equation*}
    a_{mn}=\frac{\exp\left(l_{mn}/\sqrt{d/H}\right)}{\sum_{i=1}^L \exp\left(l_{mi}/\sqrt{d/H}\right)}
\end{equation*}

\begin{figure}
    \centering
    \includegraphics[width=\linewidth]{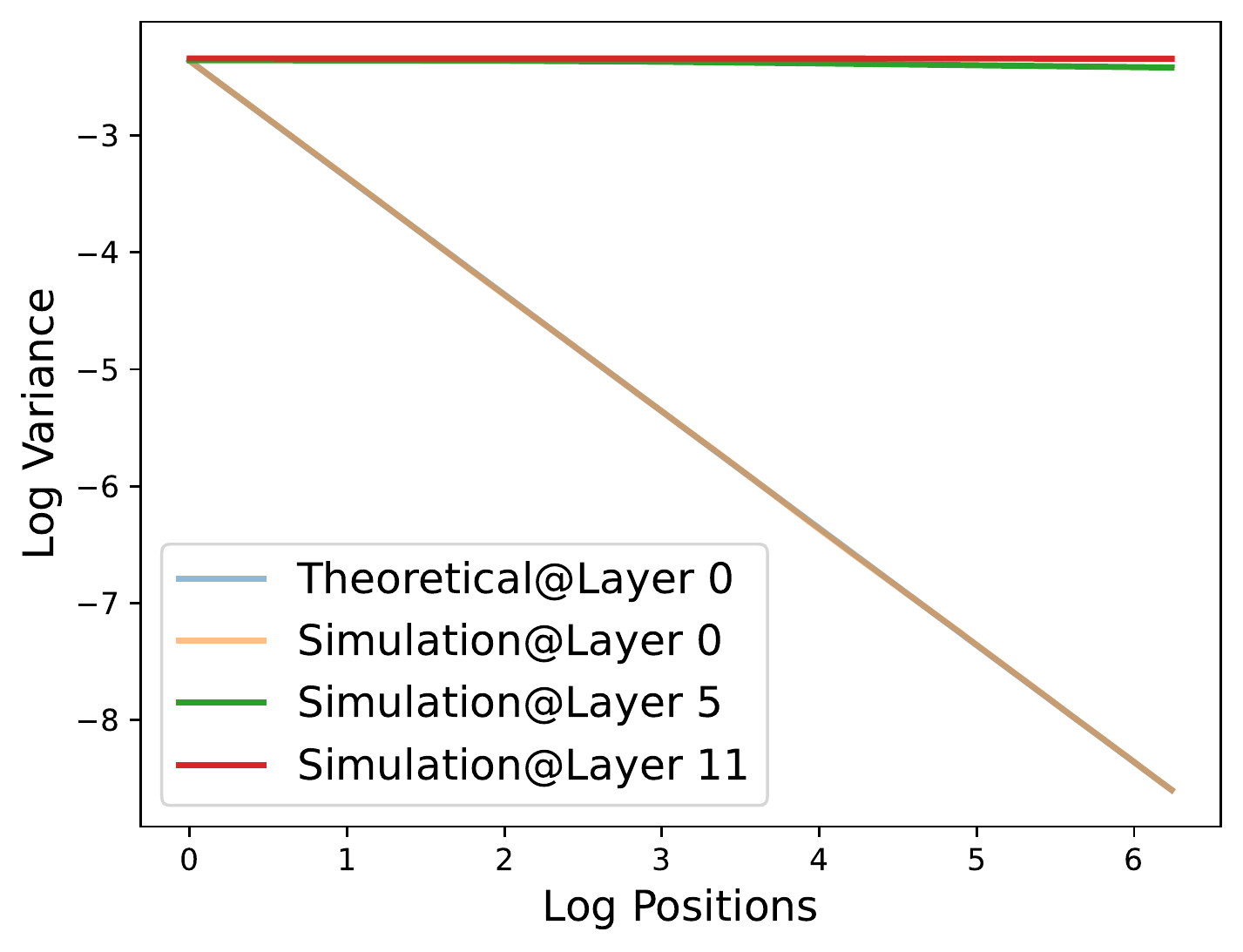}
    \caption{We plot the log positions (up to $L=512$) w.r.t their log variance under the assumption of Property~\ref{prop:lmn}. The simulation aligns with the theoretical curve posited by Lemma~\ref{lemma:cov_output} at the $0^\text{th}$ layer averaged over 500 samples.}
    \label{fig:reduced_variance}
\end{figure}

\begin{lemma}
$l_{mn}$ has zero mean and $\frac{d^3\sigma^4}{H^2}$ variance. $l_{mn}/\sqrt{d/H}$ has $\frac{d^2\sigma^4}{H}$ variance.
\label{lemma:cov_lmn}
\end{lemma}
\noindent The numerical variance of $l_{mn}/\sqrt{d/H}$ in our case is $\frac{768^2\cdot0.02^4}{12}\approx0.0079$. Lemma~\ref{lemma:cov_lmn} suggests the following approximation:
\begin{property}
When $\sigma^4 \ll \frac{H}{d^2}$, $l_{m,:}$ has small variance, making the attention weights $a_{m,:}$ almost evenly distributed among all positions.\footnote{This approximation was also used in~\citet{Xiong2020OnLN} except that they made a stronger assumption that $\bm W_q$ and $\bm W_k$ have to be initialized as zero matrices.}
\label{prop:lmn}
\end{property}
\noindent In Figure~\ref{fig:avg_attn}, we verify Property~\ref{prop:lmn} by showing that $\bm a_{mn}$ is almost evenly distributed in simulation.

\begin{figure*}
    \centering
    \includegraphics[width=0.8\linewidth]{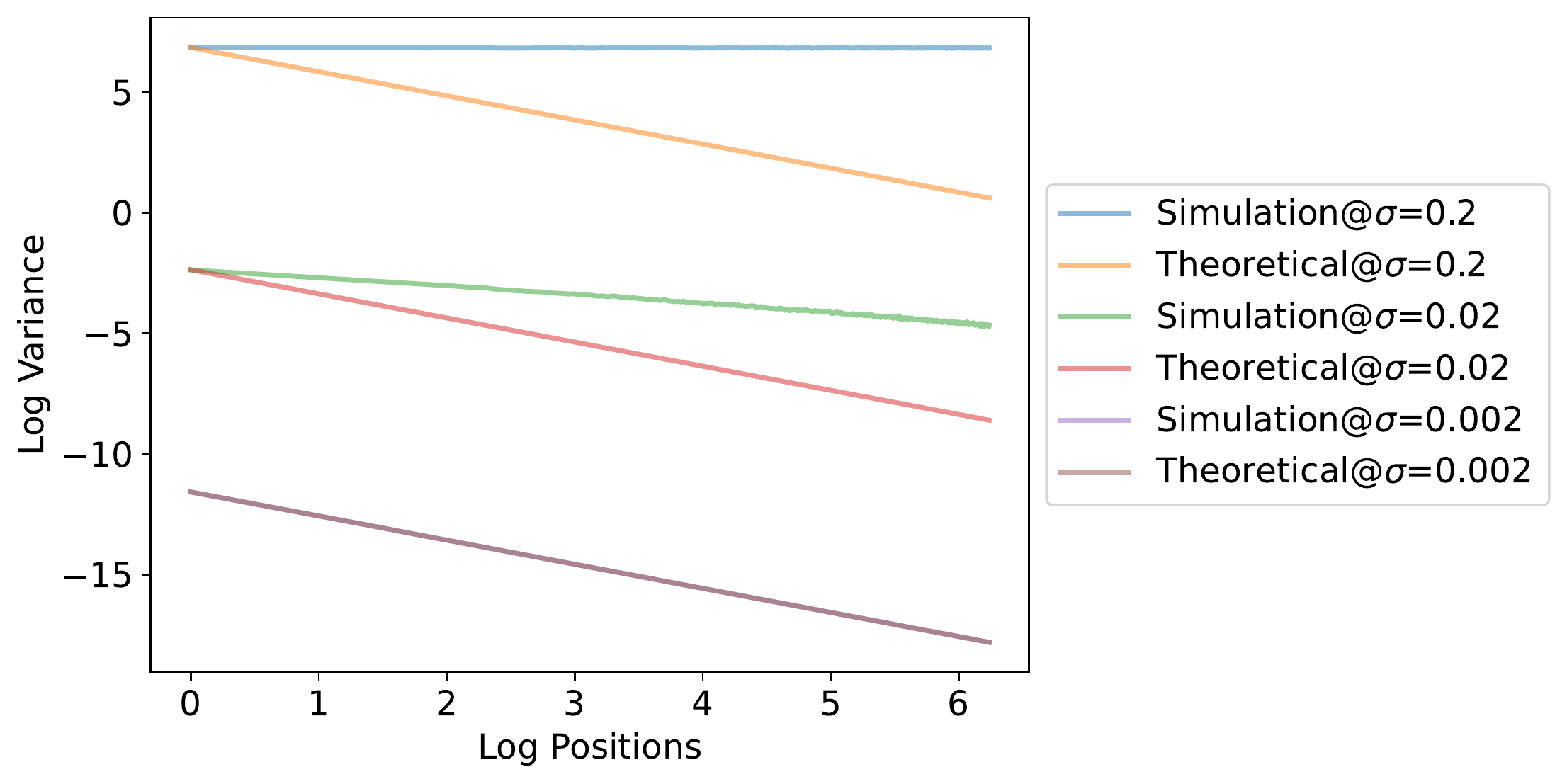}
    \caption{We vary the value of $\sigma$ and show its effect at the $0^\text{th}$ layer. As we can see, a smaller value of $\sigma$ brings Lemma~\ref{lemma:cov_output} into alignment with the corresponding simulation more closely. Note that the two lines overlap completely when $\sigma=0.002$. Average of 500 samples.}
    \label{fig:real_init_variance}
\end{figure*}

Observe that the output vector $\bm o_m$ at position $m$ is:
\begin{equation*}
    \bm o_m = \bm W_o \left(\oplus_{h=1}^H \sum_{n=1}^L a_{mn}^{(h)}\bm v_n^{(h)} \right),
\end{equation*}
where $\oplus$ denotes the concatenation of vectors from all $H$ attention heads. Assume that Property~\ref{prop:lmn} is valid and that $\bm W_o\in\mathbb{R}^{d\times d}$ has elements i.i.d sampled from $N(0,\sigma^2)$, we derive the distribution of $\bm o_m$ below.
\begin{lemma}
    $\bm o_m$ has zero mean and $\frac{d^2\sigma^4}{m} I$ covariance matrix.
    \label{lemma:cov_output}
\end{lemma}
\noindent Figure~\ref{fig:reduced_variance} is a simulation that verifies Lemma~\ref{lemma:cov_output} under the assumption of Property~\ref{prop:lmn}. We can see that~\emph{the variance of $\bm o_m$ already encodes the positional information $m$.}

\subsection{Residual Connection}
\label{sec:residual}
As denoted by the~\emph{Addition} block of Figure~\ref{fig:pre_ln}, the residual connection sets the output as $\bm y_m=\bm x_m+\bm o_m$. It allows the model to pass the first MHA output to later MHA modules as well as the final classifier. As the positional information has been passed by the residual connection, we omit the FFN part in our analysis.

\subsection{The Final Layer Normalization}
\label{sec:final_ln}
Layer normalization is an operation that might eliminate the positional information derived in Lemma~\ref{lemma:cov_output}, which happens before the MHA modules and position classifier. As mentioned in \S\ref{sec:ln},
$\text{LN}(\bm y_m)$ gives:
\begin{equation*}
    \bm y'_{mi} \approx \frac{\bm y_{mi}-\E[\bm y_{mi}]}{\sqrt{\mathbb{V}[\bm y_{mi}]}} \approx \frac{\bm x_{mi} + \bm W_o\bm W_v\frac{\sum_n^m \bm e_{ni}}{m}}{\sqrt{\sigma^2+\frac{d^2\sigma^4}{m}}},
\end{equation*}

\begin{align*}
    \mathbb{E}[\bm y_{mi}] = 0,~\mathbb{V}[\bm y_{mi}] &= \mathbb{V}[\bm x_{mi}] + \mathbb{V}[\bm o_{mi}] \\ &= \sigma^2 + \frac{d^2\sigma^4}{m}
\end{align*}

\begin{lemma}
    The variance of the $j$-th dimension of $\bm y_m$ is: \begin{equation*}
    \frac{m\sigma^2+\sum_i(\bm W_{o,j:} \bm W_{v,:i})^2}{m\sigma^2+d^2\sigma^4},
    \label{lemma:final_ln}
\end{equation*}
\end{lemma}
\noindent where $\bm W_{o,j:}\in\mathbb{R}^{1\times d}$ is the $j$-th row of $\bm W_o$.
$\bm W_{v,:i}\in\mathbb{R}^{d\times 1}$ is the $i$-th column of $\bm W_v$. As long as $\sum_i(\bm W_{o,j:} \bm W_{v,:i})^2 \neq d^2\sigma^4$, the classifier should be able to exploit the discrepancy to derive $m$.

Readers might wonder why $\bm W_{o,j:}$ and $\bm W_{v,:i}$ in the numerator cannot be treated as random variables. The reason is that we only focus on one dimension ($j$-th) at a time. This means we cannot use the law of large numbers to approximate the sample variance of $\bm y_{mj}$ as we did for the denominator.

\subsection{Relaxing the Assumptions}
We discuss possible relaxation of the assumptions used in \S\ref{sec:lemmas-for-attn}.
\paragraph{What if Property~\ref{prop:lmn} does not hold?} Or equivalently, $\sigma^4 \centernot\ll \frac{H}{d^2}$. This prompts us to vary the value of $\sigma$. In Figure~\ref{fig:real_init_variance}, we see that smaller $\sigma$ better aligns Lemma~\ref{lemma:cov_output} with the simulations, which is unsurprising as Lemma~\ref{lemma:cov_output} assumes small $\sigma$. Even when $\sigma$ is not too small (i.e., $\sigma=0.2, 0.02$), the variance still encodes the positional information as the variance of $\bm o_m$ is negatively correlated with its position $m$.

\paragraph{Other Initialization Schemes}So far we assume the weight matrices ($\bm W_q$,~$\bm W_k$,~$\bm W_v$,~$\bm W_o$) are initialized i.i.d from $N(0,\sigma^2)$. However, we can relax the assumption to i.i.d. samples from a distribution with zero mean and finite variance. This is because the proof in Appendix \ref{sec:appendix_proof} calculates the covariance. The variance calculation relies on $\E[\bm r_i \bm r_i^\top] = \sigma^2 I$ where $\bm r_i^\top$ is the i-th row vector of a weight matrix. This property holds for any distribution with zero mean and $\sigma^2$ variance.

\section{Discussions}
\paragraph{Why are the positions of later tokens in a sequence harder to be predicted in Figure 3 of~\citet{haviv2022transformer}?}
Lemma~\ref{lemma:cov_output} states the variance is inversely proportional to the position $m$, so the variance of later tokens (large $m$) plateaus, resulting in a harder numerical optimization problem. This also suggests a potential downside of removing positional embeddings: It might be challenging for the model to infer positional information of the later tokens in extremely long input sequences.
\paragraph{Why do lower layers (closer to input) give worse probing performances in both Figure~\ref{fig:mae} and~\citet{haviv2022transformer}?}
This can be explained by Figure~\ref{fig:reduced_variance}. Most of the positions at the $0^\text{th}$ layer have tiny variance ($\exp(-10)=4.5e^{-5}$), which again poses a difficult numerical optimization problem.
\paragraph{Why does BERT fail to converge without positional embeddings?}
In a BERT model~\cite{bert}, each token has access to all the other tokens, making the variance at all positions $\frac{d^2\sigma^4}{L}$. Therefore, a BERT model cannot utilize variance differences as its positional indicator.

\section{Post-Training Results}
\label{sec:real_world}
Our derivations only apply to the initial stage where the TLM and input embeddings are randomly initialized, which may not hold true after gradient updates. It is essential to verify the existence of variance properties and lemmas on a fully pre-trained TLM on OpenWebText2 (details in Appendix~\ref{sec:appendix_pretraining}).

\begin{figure}
    \centering
    \includegraphics[width=\linewidth]{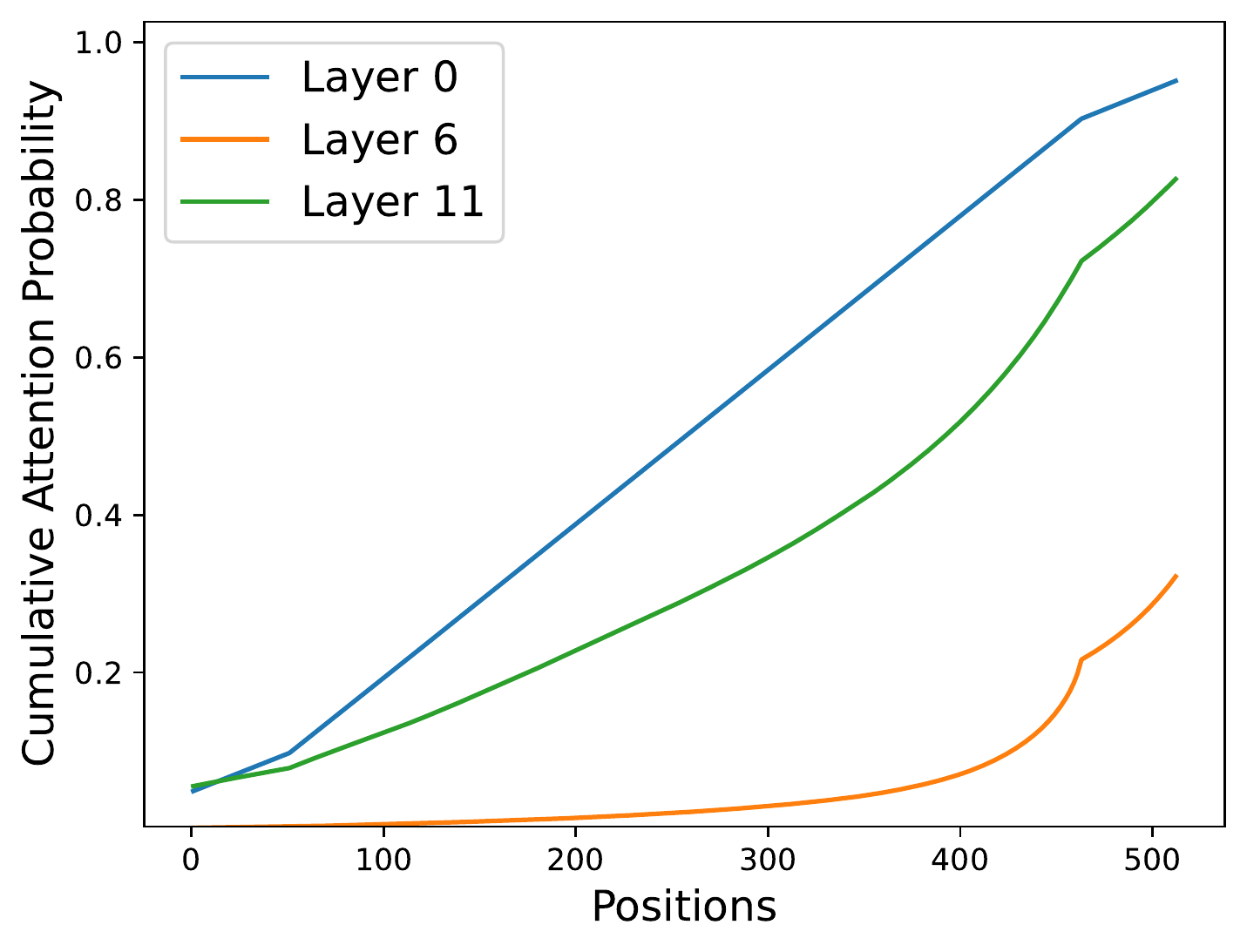}
    \caption{We plot the positions w.r.t their cumulative attention probability for $L=512$ of a pre-trained TLM. We average over all heads in a layer and 500 samples.}
    \label{fig:real_world_avg_attn}
\end{figure}

\begin{figure}
    \centering
    \includegraphics[width=\linewidth]{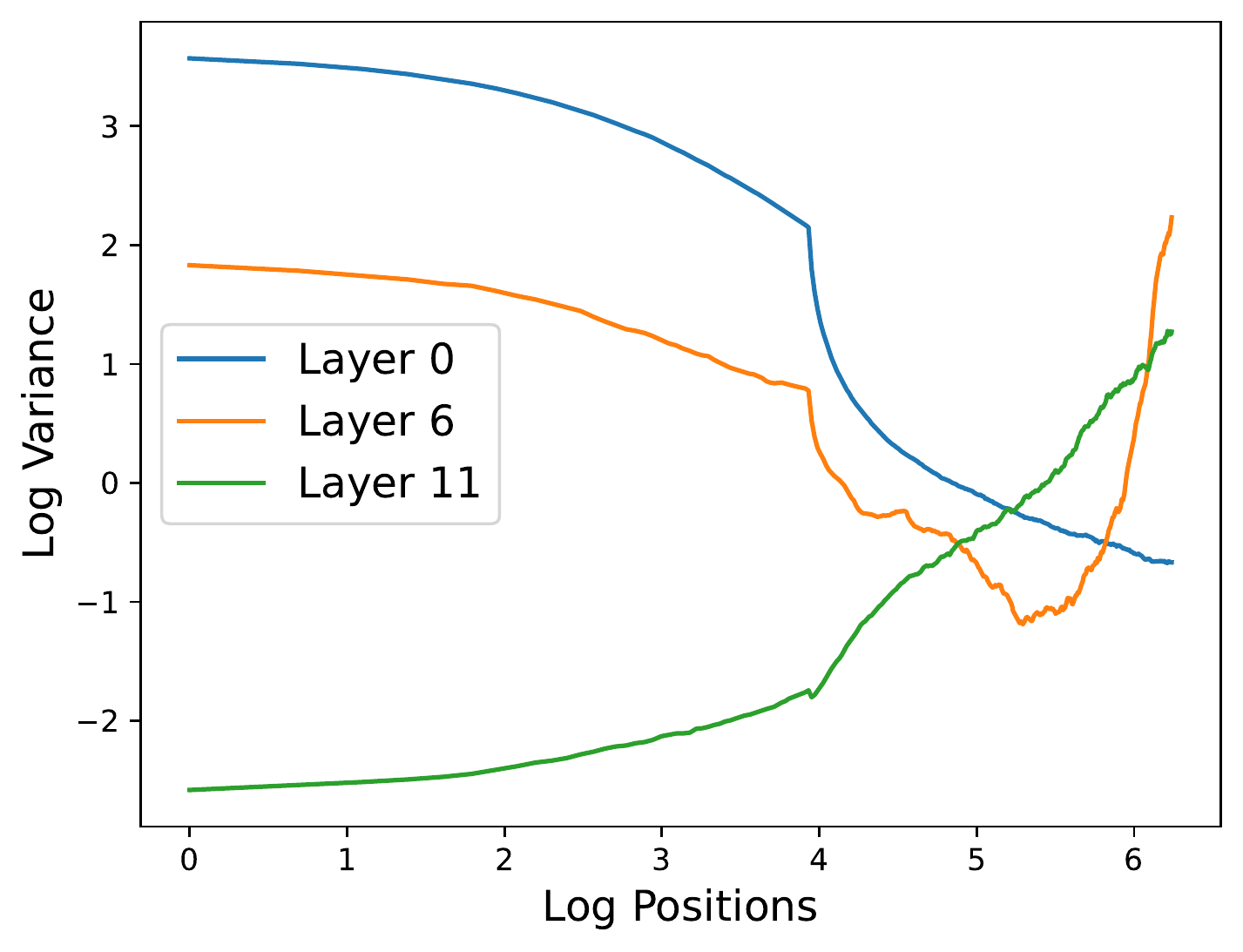}
    \caption{We plot the log positions w.r.t their log variance for $L=512$ of a pre-trained TLM. We average over 500 samples.}
    \label{fig:real_world_variance}
\end{figure}

We expect that the properties of lower layers of a pre-trained TLM should align more closely with the theoretical results for two reasons: 1) There are more steps between the lower layers and the final language modeling loss, resulting in smaller gradients and thereby fewer parameter updates, and 2) Lower layers typically encode more low-level information dependent on positional information~\cite{vulic-etal-2020-probing, de-vries-etal-2020-whats}. Figures~\ref{fig:real_world_avg_attn} and~\ref{fig:real_world_variance} demonstrate that the $0^\text{th}$ (lowest) layer exhibits highly similar cumulative attention probability and decay-with-position variance as the theoretical results. In contrast, higher layers deviate from the analyses in \S~\ref{sec:analysis}. We posit that the model learns to rely more heavily on semantic rather than positional information. This also explains why predicting positions using outputs of higher transformer layers is more challenging as demonstrated in Figure 2 of~\citet{haviv2022transformer}.

\section{Conclusion}
We mathematically analyzed a randomly initialized transformer language model without positional embeddings. We showed that the variance of the self-attention output decreases as the position increases, which serves as an indicator for positional information. We validated that, after extensive gradient updates, the low layers of a pretrained language model still exhibit highly similar variance reduction behaviors. Our results pave the way for the pretraining of more efficient and positional embedding-free transformer language models.

\section*{Limitations}
The limitations of this work mostly come from our assumptions: 1) A randomly initialized and frozen TLM, and 2) Input tokens are all different and randomly sampled. These two assumptions obviously do not hold true for human languages and pre-trained TLMs. Therefore, we attempted to empirically verify the existence of lemmas and properties on a pre-trained TLM without positional embeddings in \S\ref{sec:real_world}.

That being said, several methods could be attempted to remove these assumptions.
Firstly, we can analyze the training dynamics of a TLM to shed light on the model parameter distribution after pretraining. Secondly, Zipf's law or a simple n-gram language model could be used to quantify the degree of input token duplication in human languages. This might give us a more accurate estimate of the variance at different positions. We leave these ideas as future work. 

\section*{Ethics Statement}
Our work provides a deeper understanding of why a transformer language model can still perform well without positional embeddings, potentially enabling the application of developing future transformers that are greener and more cost-efficient.
Inappropriate usage of our technique might have negative societal impacts though. These include the ethical challenges of improper text generation and privacy issues inherent in the data collection process.
These implications apply to any natural language processing research and are not unique to this specific work.
\section*{Acknowledgment}
The authors acknowledge the support from
Boeing (2019-STU-PA-259), 
Amazon (CC ADV 00474341 2021 TR),  
NSF MRI Award 1919452, and
Princeton Research Computing.

\bibliography{anthology}
\bibliographystyle{acl_natbib}

\clearpage

\appendix
\begin{table*}[!ht]
    \centering
    \setlength{\tabcolsep}{3pt}
    \begin{tabular}{ccccc}
        \hline\hline
         \# Layers & Hidden Size & \# Attention Heads & Train Seq. Len. & \# Trainable Params.\\
         12 & 64 & 12 & 512 & ~162M\\ \hline
         Optimizer & Batch Size & Train Steps & Precision & Dataset\\
         Adam (lr 6e-4) & 32 & 50,000 & bfloat16 & OpenWebText2\\
         \hline\hline
    \end{tabular}
    \caption{\textbf{Pre-trained Model Configurations.}}
    \label{tab:model_configs}
\end{table*}

\section{Proofs}
\label{sec:appendix_proof}

The proof of Lemma~\ref{lemma:cov_v} and~\ref{lemma:cov_lmn} are head-dependent while that of Lemma~\ref{lemma:cov_output} is head-independent. For notation simplicity, at Lemma~\ref{lemma:cov_v} and~\ref{lemma:cov_lmn}, we drop the head dependency on matrices ($\bm W_q^{(h)}$, $\bm W_k^{(h)}$, $\bm W_v^{(h)}$), vectors ($\bm q_m^{(h)}$, $\bm k_m^{(h)}$, $\bm v_m^{(h)}$), and scalars ($l_{mn}^{(h)}$, $a_{mn}^{(h)}$).

\paragraph{Proof of Lemma~\ref{lemma:cov_v}}
Here, we use $\bm r_i^\top$ to denote the $i$-th row vector of $\bm W_v$.
\begin{equation*}
\begin{split}
    \text{cov}&(\bm v_{m}, \bm v_{n}) = \mathbb{E}[\bm v_m\bm v_n^\top]\\
    &=\E[\bm W_v \bm e_m \bm e_n^\top \bm W_v^\top]\\
    &=\E\left[ \begin{bmatrix}
    \bm r_1^\top \bm e_m \\ \vdots \\ \bm r_{\frac{d}{H}}^\top \bm e_m
    \end{bmatrix}\begin{bmatrix}
    \bm e_n^\top \bm r_1 & \hdots & \bm e_n^\top \bm r_{\frac{d}{H}}
    \end{bmatrix}\right]\\
    &=\left[ \E[\bm r_i^\top \bm e_m \bm e_n^\top \bm r_j] \right]_{i,j=1}^\frac{d}{H}\\
    &=\left[ \E[\tr(\bm r_j \bm r_i^\top \bm e_m \bm e_n^\top)] \right]_{i,j=1}^\frac{d}{H}\\
    &=\left[ \tr(\E[\bm r_j \bm r_i^\top] \E[\bm e_m \bm e_n^\top]) \right]_{i,j=1}^\frac{d}{H}\\
    &\overset{(*)}{=}\left[ \tr((\mathbbm{1}_{i=j}\sigma^2)\cdot I_d\cdot\mathbbm{1}_{m=n}\cdot I_d) \right]_{i,j=1}^\frac{d}{H}\\
    &=\left[ \mathbbm{1}_{i=j}\mathbbm{1}_{m=n}d\sigma^2\right]_{i,j=1}^\frac{d}{H}\\
    &=(\mathbbm{1}_{m=n}d\sigma^2)\cdot I_{d/H}
    \label{eq:cov-v}
\end{split}
\end{equation*}
$(*)$ holds because $\bm r_i$ and $\bm r_j$ are independent when $i\neq j$ (similarly for $\bm e_m$ and $\bm e_n$) and the covariance of a Gaussian random vector is an identity matrix. $I_d$ and $I_{d/H}$ denote $d\times d$ and $\frac{d}{H}\times \frac{d}{H}$ identity matrices.

\paragraph{Proof of Lemma \ref{lemma:cov_lmn}}
Here, we use $\bm r_i^\top$ to denote the $i$-th row vector of $\bm W_q$ and $\bm W_k$.
\begin{align*}
    \text{cov}&(l_{mn}, l_{mp})\\
    &=\E[(\bm e_m^\top \bm W_q^\top \bm W_k \bm e_n)(\bm e_m^\top \bm W_q^\top \bm W_k \bm e_p)^\top]\\
    &=\E[\tr(\bm e_m^\top \bm W_q^\top \bm W_k \bm e_n \bm e_p^\top \bm W_k^\top \bm W_q \bm e_m)]\\
    &=\E[\tr(\bm e_m\bm e_m^\top \bm W_q^\top \bm W_k \bm e_n \bm e_p^\top \bm W_k^\top \bm W_q )]\\
    &=\tr(\E[\bm e_m\bm e_m^\top] \E[\bm W_q^\top \bm W_k \bm e_n \bm e_p^\top \bm W_k^\top \bm W_q ])\\
    &=\E[\tr(\bm e_n \bm e_p^\top \bm W_k^\top \bm W_q \bm W_q^\top \bm W_k)]\\
    &=\tr(\E[\bm e_n\bm e_p^\top]\E[\bm W_k^\top \bm W_q \bm W_q^\top \bm W_k)]) \\
    &=(\mathbbm{1}_{n=p})\tr(\E[\bm W_q \bm W_q^\top]\E[\bm W_k \bm W_k^\top])\\
    &\overset{(*)}{=}(\mathbbm{1}_{n=p})\tr((\frac{d}{H}\sigma^2 \cdot I)(\frac{d}{H}\sigma^2 \cdot I)) \\
    &=(\mathbbm{1}_{n=p}) \frac{d^3\sigma^4}{H^2}
\end{align*}
$(*)$ holds since:
\begin{align*}
    \E[\bm W_q \bm W_q^\top]&=\E\left[ \begin{bmatrix}
    \bm r_1^\top \\ \vdots \\ \bm r_\frac{d}{H}^\top
    \end{bmatrix}\begin{bmatrix}
    \bm r_1 & \hdots & \bm r_\frac{d}{H}
    \end{bmatrix}\right]\\
    &= \left[ \E[\bm r_i^\top \bm r_j] \right]_{i,j=1}^\frac{d}{H} = \frac{d}{H}\sigma^2\cdot I
\end{align*}

\paragraph{Proof of Lemma~\ref{lemma:cov_output}}
Because $\bm W_o\in \mathbb{R}^{d\times d}$ is applied on a concatenation of vectors at all heads, we take $\bm v_i = \oplus_{h=1}^H \bm v_i^{(h)}$. $\bm v_i$ here is head-independent while $\bm v_i$ at Lemma~\ref{lemma:cov_v} is head-dependent. Here, we use $\bm r_i^\top$ to denote the $i$-th row vector of $\bm W_o$.

\begin{equation*}
\begin{split}
    \text{cov}&(\bm o_m,\bm o_m)\\
    &\overset{\text{Property}~\ref{prop:lmn}}{\approx}\E\left[ \bm W_o\frac{\sum_{i=1}^m\bm v_i}{m} \frac{\sum_{j=1}^m\bm v_j^\top}{m}\bm W_o^\top \right]\\
    &=\frac{1}{m^2}\sum_{i,j=1}^m \E[\bm W_o \bm v_i \bm v_j^\top \bm W_o^\top]\\
    &=\frac{1}{m^2}\sum_{i,j=1}^m\E\left[ \begin{bmatrix}
    \bm r_1^\top \bm v_i \\ \vdots \\ \bm r_d^\top \bm v_i
    \end{bmatrix}\begin{bmatrix}
    \bm v_j^\top \bm r_1 & \hdots & \bm v_j^\top \bm r_d
    \end{bmatrix}\right]\\
    &=\frac{1}{m^2}\sum_{i,j=1}^m\left[ \E[\bm r_k^\top \bm v_i \bm v_j^\top \bm r_l] \right]_{k,l=1}^d\\
    &=\frac{1}{m^2}\sum_{i,j=1}^m\left[ \E[\tr(\bm r_l \bm r_k^\top \bm v_i \bm v_j^\top)] \right]_{k,l=1}^d\\
    &=\frac{1}{m^2}\sum_{i,j=1}^m\left[ \tr(\E[\bm r_l \bm r_k^\top] \E[\bm v_i \bm v_j^\top]) \right]_{k,l=1}^d\\
    &\overset{(*)}{=}\frac{1}{m^2}\sum_{i,j=1}^m\left[ \tr((\mathbbm{1}_{k=l}\sigma^2)\cdot I\cdot(\mathbbm{1}_{i=j} d\sigma^2)\cdot I)\right]_{k,l=1}^d\\
    &=\frac{d^2\sigma^4}{m}I
\end{split}
\end{equation*}
$(*)$ follows from Lemma~\ref{lemma:cov_v}: because $\text{cov}(\bm v_{i}^{(h)}, \bm v_{j}^{(h)})=(\mathbbm{1}_{i=j}d\sigma^2)\cdot I_{d/H}$, a concatenation for all $h\in H$ gives $\mathbb{E}[\bm v_i\bm v_j^\top]=(\mathbbm{1}_{i=j}d\sigma^2)\cdot I_d$.

\section{Probing Experiment Details}
\label{sec:appendix_probing}
We train a randomly initialized and frozen TLM with $12$ layers, $d=768$, $H=12$, $L=512$, and $\sigma=0.02$. We use the Adam optimizer~\cite{kingma2014method} with learning rate $1e-3$ and $5000$ gradient updates. The batch size is set to $32$. We implement our model using PyTorch~\cite{NEURIPS2019_9015}.

\section{Pre-trained Transformer Language Model Details}
\label{sec:appendix_pretraining}
We use the gpt-neox library~\cite{gpt-neox-library} to train a TLM with no positional embeddings. Detailed hyperparameters are listed in Table~\ref{tab:model_configs}. The pretraining takes 5 hours on one NVIDIA A100-40GB.

\section{Scientific Artifacts}
\label{sec:artifact}
We use the gpt-neox library~\cite{gpt-neox-library} under Apache-2.0 license. OpenWebText2~\cite{pile} is released by the authors of gpt-neox. The codebase and dataset are publicly released for research purposes. The steps taken to protect privacy and anonymization are discussed in Section 6 and 7 of~\citet{pile}. The distribution and statistics of OpenWebext2 are also discussed in~\citet{pile}.
\end{document}